\pdfoutput=1

\documentclass[11pt]{article}

\usepackage[]{ACL2023}

\usepackage{times}
\usepackage{latexsym}

\usepackage[T1]{fontenc}

\usepackage[utf8]{inputenc}

\usepackage{microtype}

\usepackage{inconsolata}

\usepackage{enumitem}
\usepackage[labelformat=simple]{subcaption}

\usepackage{graphicx}
\usepackage{amsmath}
\usepackage{verbatim}
\usepackage{algorithm}
\usepackage{algpseudocode}
\usepackage{amsmath}
\usepackage{amssymb}
\usepackage{multirow}
\usepackage{makecell}
\usepackage{booktabs}
\usepackage{stfloats}
\usepackage[misc]{ifsym}
\newcommand*\samethanks[1][\value{footnote}]{\footnotemark[#1]}

\title{CASE: Aligning Coarse-to-Fine Cognition and Affection for\\Empathetic Response Generation}

\author{
    Jinfeng Zhou\textsuperscript{\rm 1,2}\thanks{\ \ Work done during internship at the CoAI Group.}\ \ \thanks{\ \ Equal contribution.}\quad
    Chujie Zheng\textsuperscript{\rm 1}\samethanks{}\quad 
    Bo Wang\textsuperscript{\rm 2}\thanks{\ \ Corresponding author.}\quad 
    Zheng Zhang\textsuperscript{\rm 1}\quad 
    Minlie Huang\textsuperscript{\rm 1} \\ 
    \small \textsuperscript{\rm 1}The CoAI Group, DCST, Institute for Artificial Intelligence, State Key Lab of Intelligent Technology and Systems, \\
    \small \textsuperscript{\rm 1}Beijing National Research Center for Information Science and Technology, Tsinghua University, Beijing 100084, China \\
    \small \textsuperscript{\rm 2}College of Intelligence and Computing, Tianjin University, Tianjin, China \\
    \small \texttt{\{jfzhou.mail, chujiezhengchn, zhangz.goal\}@gmail.com, bo\_wang@tju.edu.cn, aihuang@tsinghua.edu.cn}
}

\begin{document}

\maketitle

\begin{abstract}

Empathetic conversation is psychologically supposed to be the result of conscious alignment and interaction between the cognition and affection of empathy. 
However, existing empathetic dialogue models usually consider only the affective aspect or treat cognition and affection in isolation, which limits the capability of empathetic response generation.
In this work, we propose the CASE model for empathetic dialogue generation.
It first builds upon a commonsense cognition graph and an emotional concept graph and then aligns the user's cognition and affection at both the coarse-grained and fine-grained levels.
Through automatic and manual evaluation, we demonstrate that CASE outperforms state-of-the-art baselines of empathetic dialogues and can generate more empathetic and informative responses.\footnote{The project repository is available at \url{https://github.com/jfzhouyoo/CASE}}

\end{abstract}

\section{Introduction}

Human empathetic conversations allow both parties to understand each other's experiences and feelings \cite{keskin2014isn}, which is crucial for establishing seamless relationships \cite{zech2005talking} and is also integral to building a trustful conversational AI \cite{DBLP:journals/tois/HuangZG20,wang2021cass}. 

In social psychology, empathy consists of two aspects: cognition and affection \cite{davis1983measuring}. 
The cognitive aspect corresponding to the understanding of the user's \textit{situation and experiences} \cite{cuff2016empathy}. 
The affective aspect requires the comprehension of the user's \textit{emotional state} and his/her potential \textit{emotional reaction} \cite{elliott2018therapist}.
Although existing work of empathetic dialogue involves both aspects of empathy, there are still issues that need to be addressed. 
\textbf{First}, most work \cite{DBLP:conf/acl/RashkinSLB19, DBLP:conf/emnlp/LinMSXF19, DBLP:conf/emnlp/MajumderHPLGGMP20, DBLP:conf/coling/LiCRRTC20, li2022knowledge} considers only the affective aspect, like detecting the user's emotional state to enhance empathy expression. 
\textbf{Second}, although recent work explored both roles of cognition and affection in empathy expression \cite{DBLP:conf/acl/ZhengLCLH21, sabour2021cem}, they usually treat cognition and affection in isolation without considering their relationship.

\begin{figure}[t]
\centering
\includegraphics[width=\columnwidth]{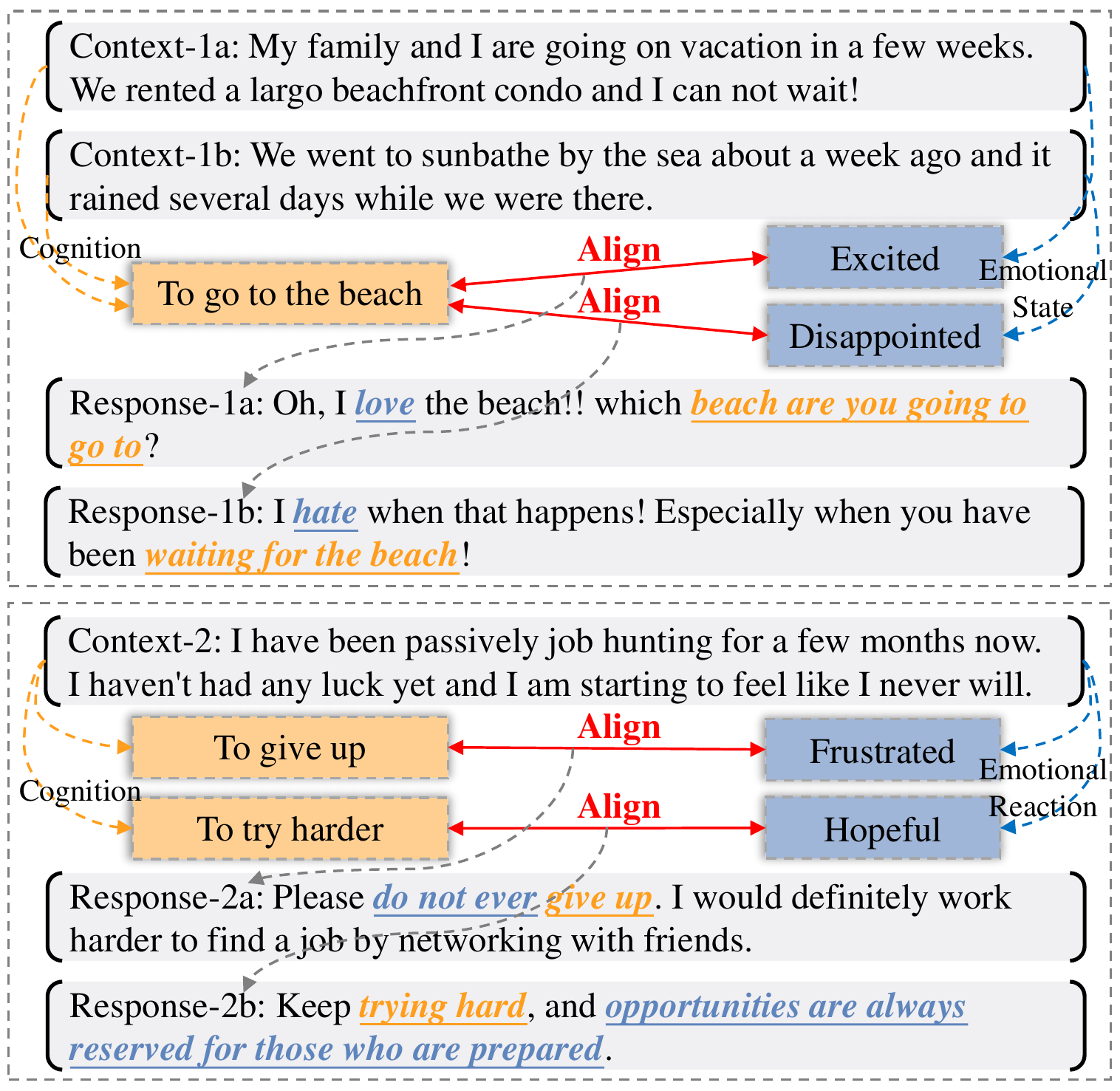} 
\caption{Examples from the \textsc{EmpatheticDialogues} dataset. The alignment of cognition and affection (i.e., emotional state and emotional reaction) leads to highly empathetic and informative expression in responses.}
\label{introcase}
\end{figure}

However, human empathetic responses often result from conscious alignment and interaction between cognition and affection of empathy \cite{westbrook2011introduction}.
For one thing, the user's overall emotional state manifested in the context suggests the user's attitude toward current situation (i.e., cognition). Thus, for the listener, aligning the user's expressed cognition to the proper emotional state is essential for an appropriate empathetic response. As in case-1 of Figure \ref{introcase}, the alignment of cognition (i.e., \textit{intent} “\textit{to go to the beach}”) with different emotional states (i.e., “\textit{excited}” vs. “\textit{disappointed}”) produces different appropriate empathetic expressions (i.e., “\textit{love}” and “\textit{which beach are you going to go to}” vs. “\textit{hate}” and “\textit{waiting for the beach}”), respectively.
For another, the user's situation drives the listener to infer the deeper specific cognitions and associate them with the underlying emotional reactions. In this way, the listener can produce a more actively empathetic response instead of only understanding and repeating the user's expressed cognition. As in case-2 of Figure \ref{introcase}, building association between inferred cognitions and emotional reactions, i.e., “\textit{to give up}” and “\textit{frustrated}” vs. “\textit{to try harder}” and “\textit{hopeful}”, yields cognitively distinct but highly empathetic responses, i.e., \textit{response-2a} vs. \textit{response-2b}.
The two cases highlight the necessity of aligning cognition and affection on both overall and specific (i.e., coarse and fine-grained) level for empathy modeling in response generation.

To this end, we align \textbf{C}ognition and \textbf{A}ffection for re\textbf{S}ponding \textbf{E}mpathetically (\textbf{CASE}) on coarse and fine-grained levels by fusing sentence-level commonsense knowledge from COMET \cite{DBLP:conf/acl/BosselutRSMCC19} and word-level concept knowledge from ConceptNet \cite{DBLP:conf/aaai/SpeerCH17}. Commonsense knowledge infers the user’s situation as cognition and infers emotional reactions to the situation, which are implied in the dialogue. 
Concept knowledge serves to extract the emotional state manifested in the dialogue. For encoding the two types of knowledge, we first construct commonsense cognition graph and emotional concept graph, where the initial independent representation of cognition and emotional concept is carefully adjusted by dialogue context adopting graph transformers. Then, we design a two-level strategy to align cognition and affection using mutual information maximization (MIM) (Appendix \ref{sec:mim}) \cite{DBLP:conf/iclr/HjelmFLGBTB19}. The coarse-grained level considers overall cognition and affection manifested in the dialogue context to align contextual cognition and contextual emotional state, which are extracted with a knowledge discernment mechanism. The fine-grained level builds the fine-grained association between cognition and affection implied in the dialogue to align each specific cognition and corresponding emotional reaction. Further, an empathy-aware decoder is devised for generating empathetic expressions.

Our contributions are summarized as follows:

(1) We devise a unified framework to model the interaction between cognition and affection for integrated empathetic response generation.

(2) We construct two heterogeneous graphs involving commonsense and concept knowledge to aid in the modeling of cognition and affection.

(3) We propose a two-level strategy to align coarse-grained and fine-grained cognition and affection adopting mutual information maximization.

(4) Extensive experiments demonstrate the superior of CASE in automatic and manual  evaluation.

\section{Related Work}

\subsection{Emotional \& Empathetic Conversation}

Emotional conversation gives the manually specified label preset as the emotion generated in the response \cite{DBLP:conf/aaai/ZhouHZZL18, DBLP:conf/cikm/0002LMGZZH19, DBLP:conf/icassp/PengHXXZS22}. Instead of giving a predefined emotion label, empathetic conversation \cite{DBLP:conf/naacl/ChenLY22, DBLP:conf/naacl/KimAKL22} involves cognitive and affective empathy \cite{davis1983measuring} and aims to fully understand the interlocutor's situation and feelings and respond empathically \cite{keskin2014isn, zheng-etal-2021-comae}. For one thing, most existing works only focus on the affective aspect of empathy and make efforts to detect contextual emotion \cite{DBLP:conf/acl/RashkinSLB19, DBLP:conf/emnlp/LinMSXF19, DBLP:conf/emnlp/MajumderHPLGGMP20, DBLP:conf/coling/LiCRRTC20, li2022knowledge} while ignoring the cognitive aspect. For another, some research utilizes commonsense as cognition to refine empathetic considerations \cite{sabour2021cem}. However, the relatively independent modeling between the two aspects (i.e., cognition and affection) violates their interrelated characteristics.

\subsection{Commonsense \& Concept Knowledge}

As a commonsense knowledge base, ATOMIC \cite{DBLP:conf/aaai/SapBABLRRSC19} focuses on inferential knowledge organized as typed \textit{if-then} relations. Six commonsense reasoning relations are defined for the person involved in an event, four of which are used  to reason commonsense cognitions of a given event, i.e., PersonX’s intent before the event (\textit{xIntent}), what PersonX need to do before the event (\textit{xNeed}), what PersonX want after the event (\textit{xWant}), and the effect of the event on PersonX (\textit{xEffect}). Each commonsense cognition is aligned with user's emotional reaction to the situation implied in the dialogue inferred by \textit{xReact} (i.e., PersonX’s reaction to the event) in our approach. To obtain inferential commonsense knowledge, we use COMET \cite{DBLP:conf/acl/BosselutRSMCC19}, a pretrained generative model, to generate rich commonsense statements.

Unlike commonsense knowledge that provides sentence-level commonsense expression, we adopt ConceptNet \cite{DBLP:conf/aaai/SpeerCH17} as concept knowledge, which provides word-level human knowledge and is widely used in various NLP tasks \cite{DBLP:conf/acl/ZhangLXL20, DBLP:conf/aaai/ZhongWLZWM21, DBLP:conf/emnlp/ZhouWHH21, DBLP:conf/coling/YangWZTZHHH22}. Following \citet{li2022knowledge}, we use NRC$\_$VAD \cite{DBLP:conf/acl/Mohammad18} to assign emotion intensity to concepts in ConceptNet (processing details are in \citet{li2022knowledge}) severed to extract the contextual emotional state manifested in the context, and align it with contextual cognition.

\begin{figure*}[t]
\centering
\includegraphics[width=0.8\textwidth]{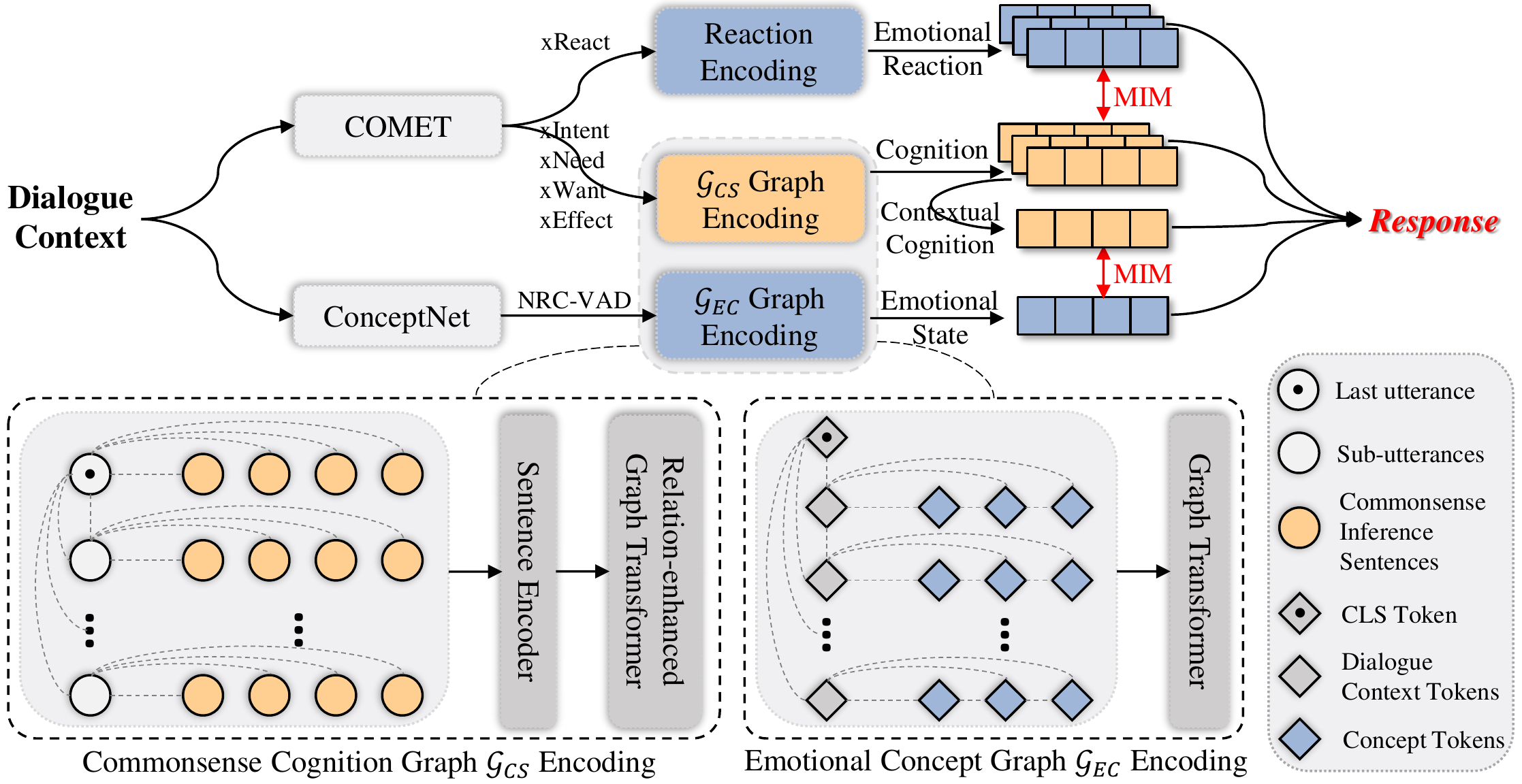} 
\caption{The architecture of the proposed CASE model.}
\label{framework}
\end{figure*}

\section{Approach}

CASE framework is in Fig. \ref{framework}.
The dialogue context $X=\left[x_{1}, \ldots, x_{N}\right]$ contains $N$ utterances, where $x_i$ denotes the $i$-th utterance.  
CASE contains three stages:
(1) The graph encoding stage constructs and encodes heterogeneous commonsense cognition graph $\mathcal{G}_{CS}$ and emotional concept graph $\mathcal{G}_{EC}$ from the dialogue context $X$. 
(2) The coarse-to-fine alignment aligns coarse-grained (between contextual cognition and contextual emotional state) and fine-grained (between each specific cognition and corresponding emotional reaction) cognition and affection adopting MIM. 
(3) The empathy-aware decoder integrates the aligned cognition and affection to generate the response $Y=\left[y_{1},y_{2},\ldots,y_{M}\right]$ with empathetic and informative expressions.

\subsection{Graph Encoding}

\paragraph{Commonsense Cognition Graph Construction}

Given the last utterance $x_{N}$ of the dialogue context $X$, we segment it into the sub-utterances $U=\left[u_0, u_{1}, u_{2}, \ldots, u_{t}\right]$, where we prepend the whole $x_{N}$ as $u_0$ for maintaining the global information of $x_{N}$.
We use COMET to infer $l$ commonsense cognition knowledge $K_{i}^{r}=[k_{i, 1}^{r}, k_{i, 2}^{r}, \ldots, k_{i, l}^{r}]$ for each $u_{i} \in U$, where $r$ is one of the four commonsense relations $\mathcal{R}=\left\{\text{xIntent},\text{xNeed},\text{xWant},\text{xEffect}\right\}$, similar to \citet{sabour2021cem}.
The idea is that human responses tend to inherit the above and transfer the topic. There are differences in the topic and connotation of different sub-utterances affecting the listeners’ concerns when responding empathetically. 

For constructing the heterogeneous commonsense cognition graph $\mathcal{G}_{CS}$, we use the utterance set $U$ and the commonsense cognition knowledge set $K_{C S}=\bigcup_{i=0}^{t} \bigcup_{r \in \mathcal{R}} K_{i}^{r}$ as vertices, i.e., vertex set $V_{C S}=U \cup K_{C S}$. 
There are seven relations of undirected edges that connect vertices. 
(1) The \textit{self-loop} relation for each vertex $v_{i}^{C S} \in V_{C S}$.
(2) The \textit{global} relation between the whole $x_N$ (i.e., $u_0$) and its sub-utterances $u_{i} (i\ge 1)$. 
(3) The \textit{temporary} relation between any two successive sub-utterances $u_j$ and $u_{j+1}$. 
(4) The four commonsense relations, i.e., \textit{xIntent}, \textit{xNeed}, \textit{xWant}, \textit{xEffect}, between the utterance $u_{i} \in U$ and the corresponding $K_{i}^{r}$. 

We use a Transformer-based sentence encoder (cognition encoder) to first encode the vertices $V_{CS}$ of the graph $\mathcal{G}_{CS}$. 
For each $v_{i}^{C S} \in V_{CS}$, we prepend with a special token $\left[CLS\right]$. 
Following \citet{DBLP:conf/naacl/DevlinCLT19}, we collect the $[CLS]$ representation as the initial embedding matrix for $\mathcal{G}_{CS}$.

\paragraph{Emotional Concept Graph Construction}

We concatenate the utterances in the dialogue context $X$ to obtain the token set $T$, i.e., $T=x_{1} \oplus \ldots \oplus x_{N}=[w_1, \dots, w_n]$, where $n$ is the number of all the tokens in the utterances in $X$. 
Following \citet{li2022knowledge}, we use ConceptNet to infer the related concepts for each token $w_{i} \in T$, among which only the the top $N^{\prime}$ emotional concepts (according to the emotion intensity $\omega(c)$) are used for constructing $\mathcal{G}_{EC}$. 
Subsequently, the vertices $V_{EC}$ in the heterogeneous emotional concept graph $\mathcal{G}_{EC}$ contains a $[CLS]$ token, the dialogue context tokens $T$, and the above obtained emotional concepts. 
There are four relations of undirected edges that connect vertices. 
(1) The \textit{self-loop} relation for each vertex $v_{i}^{EC} \in V_{EC}$.
(2) The \textit{global} relation between the $[CLS]$ token and other ones. 
(3) The \textit{temporary} relation between any two successive tokens.
(4) The \textit{emotional concept} relation among a token and its related emotional concepts. 

We initialize the vertex embedding for $\mathcal{G}_{EC}$ by summing up the token embedding, the positional embedding, and the type embedding for each vertex (signaling whether it is a emotional concept or not).

\paragraph{Graph Encoder}

Given the commonsense cognition graph $\mathcal{G}_{CS}$, to capture the semantic relationship between vertices, we adopt the Relation-Enhanced Graph Transformer \cite{DBLP:conf/sigir/LiZWYW21} for graph encoding.
It employs a relation-enhanced multi-head attention mechanism (MHA) to encode vertex embedding $\hat{\boldsymbol{v}}_{v_{i}}$ for vertex $v_{i}$ (we omit the superscripts $CS$ for simplicity) as:
\begin{equation}
\hat{\boldsymbol{v}}_{v_{i}}=M H A_{v_{k} \in V_{C S}}\left(\boldsymbol{q}_{v_{i}}, \boldsymbol{k}_{v_{k}}, \boldsymbol{v}_{v_{k}}\right),
\end{equation}
where the semantic relations between vertices are injected into the query and key vectors:
\begin{equation}
\begin{aligned}
\boldsymbol{q}_{v_{i}} =\boldsymbol{v}_{v_{i}}+\boldsymbol{l}_{v_{i} \rightarrow v_{k}},\ 
\boldsymbol{k}_{v_{k}} =\boldsymbol{v}_{v_{k}}+\boldsymbol{l}_{v_{k} \rightarrow v_{i}},
\label{eq:relation}
\end{aligned}
\end{equation}
where $\boldsymbol{l}_{v_{i} \rightarrow v_{k}}$ and $\boldsymbol{l}_{v_{k} \rightarrow v_{i}}$ are learnable relation embeddings between vertices $v_{i}$ and $v_{k}$.
The self-attention is subsequently followed by a residual connection and a feed-forward layer, as done in the standard Transformer encoder \cite{DBLP:conf/nips/VaswaniSPUJGKP17}. 
Finally, we obtain the commonsense cognition embedding $\boldsymbol{cs}_i$ for each $v_{i}^{C S} \in V_{C S}$.

To encode the emotional concept graph $\mathcal{G}_{EC}$, we adopt a vanilla Graph Transformer (i.e., omitting the relation enhancement part in the above Graph Transformer).
By superimposing the emotion intensity of each token, we obtain the emotional concept embedding $\boldsymbol{ec}_i$ for each $v_{i}^{EC} \in V_{EC}$.

\subsection{Coarse-to-Fine Alignment}

\paragraph{Context Encoding}

Following previous works \cite{DBLP:conf/emnlp/MajumderHPLGGMP20, sabour2021cem}, we concatenate all the utterances in the dialogue context $X$ and prepend with a $[CLS]$ token: $[C L S] \oplus x_{1} \oplus \ldots \oplus x_{N}$.
This sequence is fed into a standard Transformer encoder (context encoder) to obtain the representation $\boldsymbol{S}_{X}$ of the dialogue context.
We denote the representation of $[CLS]$ as $\boldsymbol{s}_{X}$.

\paragraph{Coarse-grained Alignment}

To reproduce the interaction of cognition and affection manifested in the dialogue context, we align contextual cognition and contextual emotional state at an overall level. They are separately acquired by cognitive and emotional knowledge discernment mechanisms, which select golden-like knowledge guided by response.

To obtain the contextual cognitive representation $\boldsymbol{r}_{cog}$, the knowledge discernment calculates the prior cognitive distribution $P_{C S}\left(\boldsymbol{cs}_{i} \mid X \right)$ over the commonsense cognition knowledge (that is, only $K_{CS}$ rather than all the vertices $V_{CS}$ in $\mathcal{G}_{CS}$, and we thus use $1\le i \le |K_{C S}|$ for simplicity):
\begin{align}
\boldsymbol{r}_{cog} =&\sum_{i=1}^{\left|K_{C S}\right|} P_{C S}\left(\boldsymbol{cs}_{i} \mid X \right) \cdot \boldsymbol{cs}_{i}, \\
P_{C S}\left(\boldsymbol{cs}_{i} \mid X \right) &=\mathrm{softmax}_i(\boldsymbol{cs}_{i}^T \varphi_{CS}\left(\boldsymbol{s}_{X}\right)),
\end{align}
where $\varphi_{CS}(\cdot)$ is a MLP layer activated by $\tanh$.
Similarly, we calculate the prior emotional distribution $P_{EC}\left(\boldsymbol{ec}_{i} \mid X \right)$ ($1\le i \le |V_{EC}|$) and obtain the contextual emotional representation $\boldsymbol{r}_{emo}$.

During training, we use the ground truth response $Y$ to guide the learning of knowledge discernment mechanisms.
We feed $Y$ into the cognition encoder (used for initializing the embeddings of $\mathcal{G}_{CS}$ above) and the context encoder to get the hidden states $\boldsymbol{S}_{Y}^{cog}$ and $\boldsymbol{S}_{Y}^{ctx}$, where the $[CLS]$ representations are $\boldsymbol{s}_{Y}^{ctx}$ and $\boldsymbol{s}_{Y}^{cog}$ respectively.
The posterior cognitive distribution $P_{C S}\left(\boldsymbol{cs}_{i} \mid Y \right)$ and the emotional one $P_{EC}\left(\boldsymbol{ec}_{i} \mid Y \right)$ are calculated as:
\begin{align}
P_{C S}\left(\boldsymbol{cs}_{i} \mid Y \right) &=\mathrm{softmax}_i\left(\boldsymbol{cs}_{i}^T \boldsymbol{s}_{Y}^{cog}\right), \\
P_{EC}\left(\boldsymbol{ec}_{i} \mid Y \right) &=\mathrm{softmax}_i\left(\boldsymbol{ec}_{i}^T \boldsymbol{s}_{Y}^{ctx}\right).
\end{align}
We then optimize the KL divergence between the prior and posterior distributions during training:
\begin{align}
&L_{K L}=L_{K L}^{C S} + L_{K L}^{EC}, \\
L_{K L}^{C S}&=\sum_{i=1}^{\left|K_{C S}\right|} P_{C S}\left(\boldsymbol{cs}_{i} \mid Y\right) \cdot \log \frac{P_{C S}\left(\boldsymbol{cs}_{i} \mid Y\right)}{P_{C S}\left(\boldsymbol{cs}_{i} \mid X\right)}, \nonumber \\
L_{K L}^{EC}&=\sum_{i=1}^{\left|V_{EC}\right|} P_{C S}\left(\boldsymbol{ec}_{i} \mid Y\right) \cdot \log \frac{P_{EC}\left(\boldsymbol{ec}_{i} \mid Y\right)}{P_{EC}\left(\boldsymbol{ec}_{i} \mid X\right)}. \nonumber
\end{align}

To further ensure the accuracy of discerned knowledge, similar to \citet{DBLP:conf/aaai/BaiYLWL21}, we employ the BOW loss to force the relevancy between cognitive / emotional knowledge and the target response. 
The BOW loss $L_{BOW}$ is defined as:
\begin{equation}
L_{B O W}=-\frac{1}{|B|}\sum_{y_{t} \in B} \log \eta(y_{t} \mid \boldsymbol{r}_{cog}', \boldsymbol{r}_{e m o}'),
\end{equation}
where $\eta(\cdot)$ is a MLP layer followed by $\mathrm{softmax}$ and the output dimension is the vocabulary size, $B$ denotes the word bags of $Y$, $\boldsymbol{r}_{cog}'=\sum_{i=1}^{\left|K_{C S}\right|} P_{C S}\left(\boldsymbol{cs}_{i} \mid Y\right) \cdot \boldsymbol{cs}_{i}$, and $\boldsymbol{r}_{e m o}'=\sum_{i=1}^{\left|V_{E C}\right|} P_{E C}\left(\boldsymbol{ec}_{i} \mid Y\right) \cdot \boldsymbol{ec}_{i}$. 

Finally, we align the coarse-grained representations of the contextual cognition $\boldsymbol{r}_{cog}$ and the contextual emotional state $\boldsymbol{r}_{emo}$ using mutual information maximization (MIM). 
Specifically, we adopt the binary cross-entropy (BCE) loss $L_{coarse}$ as the mutual information estimator that maximizes the mutual information between $\boldsymbol{r}_{cog}$ and $\boldsymbol{r}_{emo}$:
\begin{align}
L_{coarse}&\ =2 f_{coarse}\left(\boldsymbol{r}_{cog}, \boldsymbol{r}_{e m o}\right) \nonumber \\
&-\log \sum_{\widetilde{\boldsymbol{r}}_{e m o}} \exp (f_{coarse}(\boldsymbol{r}_{cog}, \widetilde{\boldsymbol{r}}_{e m o})) \nonumber \\
&- \log \sum_{\widetilde{\boldsymbol{r}}_{cog}} \exp (f_{coarse}(\widetilde{\boldsymbol{r}}_{cog}, \boldsymbol{r}_{e m o})), 
\end{align}
where $\widetilde{\boldsymbol{r}}_{e m o}$ and $\widetilde{\boldsymbol{r}}_{cog}$ are the encoded negative samples. 
$f_{coarse}(\cdot,\cdot)$ is a scoring function implemented with a bilinear layer activated by $\mathrm{sigmoid}$ function:
\begin{equation}
f_{coarse}\left(\boldsymbol{a}, \boldsymbol{b}\right)=\sigma\left(\boldsymbol{a}^{T} \boldsymbol{W}_{coarse} \boldsymbol{b}\right).
\end{equation}

\paragraph{Fine-grained Alignment}

To simulate the interaction of fine-grained cognition and affection implied in the dialogue during human express empathy, the fine-grained alignment builds the fine-grained association between each inferred specific cognition and corresponding emotional reaction.

For each $u_{i} \in U$, we infer the commonsense knowledge about emotional reaction $K_{i}^\text{xReact}=\left[k_{i, 1}^\text{xReact}, \ldots, k_{i, l}^\text{xReact}\right]$ using COMET, which is regarded as the user’s possible emotional reaction to the current cognitive situation. 
Since $k_{i,j}^\text{xReact} \in K_{i}^\text{xReact}$ is usually an emotion word (e.g., happy, sad), we concatenate $K_{i}^\text{xReact}$ and feed it into the Transformer-based encoder (reaction encoder) to get the representation of the emotional reaction $\boldsymbol{H}_{i}^{er}$.
Similar to \cite{DBLP:conf/emnlp/MajumderHPLGGMP20} and \cite{sabour2021cem}, we use the average pooling to represent the reaction sequence, i.e., $\boldsymbol{h}_{i}^{e r}=\mathrm{Average}\left(\boldsymbol{H}_{i}^{e r}\right)$. 
To avoid over-alignment of out-of-context emotional reaction with cognition, we inject contextual information into the representation of reaction. We first connect $\boldsymbol{h}_{i}^{er}$ with the context representation $\boldsymbol{S}_{X}$ at the token level, i.e., ${\boldsymbol{S}_{i}^{e r}}[j]=\boldsymbol{S}_{X}[j] \oplus \boldsymbol{h}_{i}^{e r}$. 
Then another Transformer-based encoder takes ${\boldsymbol{S}_{i}^{e r}}$ as input and output the fused information ${\boldsymbol{S}_{i}^{e r}}'$.
We take the hidden representation of $[CLS]$ in ${\boldsymbol{S}_{i}^{e r}}'$ as the emotional reaction representation $\boldsymbol{er}_{i}$ of $u_{i}$.

Finally, we build the association between the inferred specific cognition $\{\bigcup_{j=1}^{l} \boldsymbol{cs}_{i, j}^{r}\}$ from $u_{i}$ for $r \in \mathcal{R}=\left\{\text{xIntent},\text{xNeed},\text{xWant},\text{xEffect}\right\}$ and the emotional reaction $\boldsymbol{er}_{i}$ using MIM. 
Recall that $\left\{\bigcup_{i=0}^{t} \bigcup_{r \in \mathcal{R}} \bigcup_{j=1}^{l} \boldsymbol{cs}_{i, j}^{r}\right\}$ exactly correspond to the commonsense cognition knowlege set $K_{CS}$.
The fine-grained BCE Loss $L_{fine}$ is defined as:
\begin{align}
L_{fine}&\ =\sum_{i=0}^{t}\sum_{r \in \mathcal{R}}\sum_{j=1}^l \Big[ 2 f_{fine}\left(\boldsymbol{cs}_{i, j}^{r}, \boldsymbol{er}_{i}\right) \nonumber \\
&-\log \sum_{\widetilde{\boldsymbol{er}}_{i}} \exp \left(f_{fine}\left(\boldsymbol{cs}_{i, j}^{r}, \widetilde{\boldsymbol{er}}_{i}\right)\right) \nonumber \\
&- \log \sum_{\widetilde{\boldsymbol{cs}}_{i, j}^{r}} \exp \left(f_{fine}\left(\widetilde{\boldsymbol{cs}}_{i, j}^{r}, \boldsymbol{er}_{i}\right)\right) \Big], 
\end{align}
where $\widetilde{\boldsymbol{er}}_{i}$ and $\widetilde{\boldsymbol{cs}}_{i, j}^{r}$ are the encoded negative samples. 
$f_{fine}(\cdot,\cdot)$ is implemented as:
\begin{equation}
f_{fine}\left(\boldsymbol{a}, \boldsymbol{b}\right)=\sigma\left(\boldsymbol{a}^{T} \boldsymbol{W}_{fine} \boldsymbol{b}\right).
\end{equation}

Altogether, the coarse-to-fine alignment module can be jointly optimized by $L_{align}$ loss:
\begin{equation}
L_{align}=L_{B O W}+L_{K L}+L_{coarse}+\alpha L_{fine},
\end{equation}
where $\alpha$ is a hyper-parameter.

\paragraph{Emotion Prediction}

We fuse the contextual emotional state and emotional reaction to distill the affective representation, where we use $\boldsymbol{er}_{0}$ as the distillation signal of emotional reaction. 
This is because $\boldsymbol{er}_{0}$ is derived from the speaker’s last utterance and represents the overall emotional reaction. 
A gating mechanism is designed to capture affective representation $\boldsymbol{r}_{aff}$:
\begin{align}
\boldsymbol{r}_{a f f} &=\mu \cdot \boldsymbol{r}_{e m o}+(1-\mu) \cdot \boldsymbol{er}_{0}, \\
\mu &=\sigma\left(\boldsymbol{w}^T_{a f f}\left[\boldsymbol{r}_{e m o} ; \boldsymbol{er}_{0}\right]\right).
\end{align}
We project $\boldsymbol{r}_{a f f}$ to predict the user's emotion $e$:
\begin{equation}
P_{e m o}(e)=\mathrm{softmax}\left(\boldsymbol{W}_{e m o} \boldsymbol{r}_{a f f}\right),
\end{equation}
which is supervised by the ground truth emotion label $e^{*}$ using the cross-entropy loss:
\begin{equation}
L_{e m o}=-\log P_{e m o}\left(e^{*}\right).
\end{equation}

\subsection{Empathy-aware Response Generation}

We employ a Transformer-based decoder to generate the response. To improve empathy perception in response generation, we devise two strategies to fuse the two parts of empathy (i.e., cognition and affection). First, we concatenate the cognitive and affective signals $\boldsymbol{r}_{cog}$ and $\boldsymbol{r}_{aff}$ with the dialogue context representation $\boldsymbol{S}_{X}$ at the token level, which is then processed by a MLP layer activated by $ReLU$ to integrate cognition and affection into the dialogue context:
\begin{equation}
\boldsymbol{S}_{X}'[i]=MLP\left(\boldsymbol{S}_{X}[i] \oplus \boldsymbol{r}_{cog} \oplus \boldsymbol{r}_{aff}\right).
\end{equation}

Second, we modify the original Transformer decoder layer by adding two new cross-attention to integrate commonsense cognition knowledge $\boldsymbol{K}_{CS}=\{ \boldsymbol{cs}_i \}_{i=1}^{|K_{CS}|}$ and emotional concept knowledge $\boldsymbol{K}_{EC}=\{ \boldsymbol{ec}_i \}_{i=1}^{|V_{EC}|}$, which are inserted between the self-attention and cross-attention for $\boldsymbol{S}_{X}'$.
The decoder then predicts the next token $y_{m}$ given the previously decoded tokens $y_{<m}$, as done in the standard Transformer decoder.
We use the negative log-likelihood loss $L_{gen}$ to optimize the decoder:
\begin{equation}
L_{g e n}=-\sum_{m=1}^{M} \log P\left(y_{m} \mid X, \mathcal{G}_{C S}, \mathcal{G}_{E C}, y_{<m}\right).
\end{equation}

Finally, we jointly optimize the alignment loss, emotion prediction loss, generation loss, and diversity loss proposed by \citet{sabour2021cem} as: $L=\gamma_{1} L_{align}+\gamma_{2} L_{e m o}+\gamma_{3} L_{g e n}+\gamma_{4} L_{d i v}$, where $\gamma_{1}$, $\gamma_{2}$, $\gamma_{3}$ and $\gamma_{4}$ are hyper-parameters.

\section{Experiments}

\subsection{Experimental Setup}

\paragraph{Dataset}

The experiments are conducted on the widely used \textsc{EmpatheticDialogues} \cite{DBLP:conf/acl/RashkinSLB19} dataset, comprising 25k open domain conversations. In a conversation, the speaker confides personal experiences, and the listener infers the situation and emotion of the speaker and responds empathetically. Following \citet{DBLP:conf/acl/RashkinSLB19}, we split the train/valid/test set by 8:1:1.

\paragraph{Baselines}

(1) \textit{Transformer} \cite{DBLP:conf/nips/VaswaniSPUJGKP17}: A vanilla Transformer-based response generation model.
(2) \textit{Multi-TRS} \cite{DBLP:conf/acl/RashkinSLB19}: A multi-task Transformer model that jointly optimizes response generation and emotion prediction. 
(3) \textit{MoEL} \cite{DBLP:conf/emnlp/LinMSXF19}: An empathy dialogue model that combines the output of multiple decoders for generating.
(4) \textit{MIME} \cite{DBLP:conf/emnlp/MajumderHPLGGMP20}: An empathy dialogue model that mimics the user's emotion for responding.
(5) \textit{EmpDG} \cite{DBLP:conf/coling/LiCRRTC20}: An empathy dialogue generator that utilizes multi-resolution user emotions and feedback.
(6) \textit{KEMP} \cite{li2022knowledge}: A knowledge-aware empathy dialogue model that only uses concept knowledge.
(7) \textit{CEM} \cite{sabour2021cem}: A commonsense-aware empathetic chatting machine that only exploits commonsense knowledge.

\paragraph{Implementation Details}

We implemented all models with Pytorch. We initialize the word embeddings with pretrained GloVE word vectors \cite{DBLP:conf/emnlp/PenningtonSM14}. The dimensionality of embeddings is set to 300 for all corresponding modules. We set hyper-parameters $l=5$, $N^{\prime}=10$, $\alpha=0.2$, $\gamma_{1}=\gamma_{2}=\gamma_{3}=1$ and $\gamma_{4}=1.5$. We use Adam optimizer \cite{DBLP:journals/corr/KingmaB14} with $\beta_{1}=0.9$ and $\beta_{2}=0.98$. The batch size is 16 and early stopping is adopted. The initial learning rate is set to 0.0001 and we varied it during training following \citet{DBLP:conf/nips/VaswaniSPUJGKP17}. The maximum decoding step is set to 30 during inference. All models are trained on a GPU-P100 machine. The training process of CASE is split into two phases. We first minimize $L_{BOW}$ for pretraining knowledge discernment mechanisms, and then minimize $L$ for training overall model.

\begin{table}[t]
\centering
\resizebox{.9\columnwidth}{!}{
\begin{tabular}{l c c c c}
\hline
    Models & PPL & Dist-1 & Dist-2 & Acc \\
\hline
    Transformer & 37.65 & 0.47 & 2.05 & - \\
    Multi-TRS & 37.45 & 0.51 & 2.12 & 0.347 \\
    MoEL & 38.35 & 0.44 & 2.10 & 0.322 \\
    MIME & 37.33 & 0.41 & 1.62 & 0.296 \\
    EmpDG & 37.77 & 0.53 & 2.26 & 0.314 \\
    KEMP & 37.32 & 0.55 & 2.31 & 0.341 \\
    CEM & 36.86 & 0.64 & 2.84 & 0.373 \\
\hline
    \textbf{CASE} & \textbf{35.37} & \textbf{0.74} & \textbf{4.01} & \textbf{0.402} \\
\hline
\end{tabular}}
\caption{Results of automatic evaluation.}
\label{automatic}
\end{table}

\subsection{Automatic Evaluation}

In the model's generation evaluation, we adopt the widely used Perplexity (\textbf{PPL}) and Distinct-1/2 (\textbf{Dist-1/2}) \cite{DBLP:conf/naacl/LiGBGD16}. Perplexity evaluates the general generation quality of a model. Distinct-1/2 evaluates the generated diversity by measuring the ratio of unique unigrams/bigrams in the response. In the model's emotion classification evaluation, we measure the accuracy (\textbf{Acc}) of emotion prediction. Following KEMP and CEM, we do not report word overlap-based automatic metrics \cite{DBLP:conf/emnlp/LiuLSNCP16}, e.g., BLEU \cite{DBLP:conf/acl/PapineniRWZ02}.

In Table \ref{automatic}, our model outperforms all baselines and achieves a significant improvement on all metrics. \textbf{First}, our model achieves about 4.0\% reduction on PPL compared to the best baseline, which shows that CASE is more likely to generate ground truth responses. \textbf{Second}, our model achieves 15.6\% and 41.2\% improvement on Dist-1/2 compared to CEM, which indicates the superiority of CASE in generating informative responses at the unigrams and bigrams level. This is attributed to the coarse-to-fine alignment that allows CASE to inject more informative commonsense cognition on the premise of ensuring the perplexity of the generated response. \textbf{Third}, our model achieves about 17.9\% and 7.8\% improvement in prediction accuracy compared to KEMP and CEM, respectively. This verifies that CASE considers both aspects of affection (i.e., contextual emotional state and emotional reaction) more effectively than focusing only on a single aspect as KEMP and CEM.

\begin{table}[t]
\centering
\resizebox{.95\columnwidth}{!}{
\begin{tabular}{l c c c c}
\hline
    Models & PPL & Dist-1 & Dist-2 & Acc \\
\hline
    \textbf{CASE} & \textbf{35.37} & \textbf{0.74} & \textbf{4.01} & \textbf{0.402} \\
\hline
    w/o Graph & 36.10 & 0.68 & 3.50 & 0.280 \\
    w/o Align & 35.75 & 0.65 & 3.34 & 0.369 \\
\hline
    w/o CSGraph & 35.51 & 0.64 & 3.18 & 0.375 \\
    w/o ECGraph & 36.24 & 0.72 & 3.94 & 0.329 \\
    w/o CGAlign & 35.67 & 0.68 & 3.60 & 0.378 \\
    w/o FGAlign & 35.55 & 0.67 & 3.43 & 0.370 \\
\hline
\end{tabular}}
\caption{Results of overall-to-part ablation study.}
\label{ablation}
\end{table}

\subsection{Overall-to-Part Ablation Study}

We conduct an overall-to-part ablation study in Table \ref{ablation}.
In the overall ablation, \textbf{first}, we remove the commonsense cognition graph and emotional concept graph, called “w/o Graph”. The emotion prediction accuracy decreases significantly, which indicates that the two heterogeneous graphs make remarkable contribution to detecting emotion. 
\textbf{Second}, we remove the coarse-to-fine alignment, called “w/o Align”. The diversity of generation decreases significantly and emotion prediction accuracy drops distinctly. It supports our motivation that the alignment of cognition and affection leads to informative and highly empathetic expression.

In the part ablation, \textbf{first}, we remove two graphs, called “w/o CSGraph” and “w/o ECGraph”, respectively. From the results, we find that the contribution of the commonsense cognition graph is mainly to improve the diversity of generation (i.e., Dist-1/2), while the role of the emotional concept graph is mainly located in the recognition of emotion (i.e., Acc). This also supports our constructed motivation. 
\textbf{Second}, we remove coarse-grained and fine-grained alignments, called “w/o CGAlign” and “w/o FGAlign”, respectively. We observe that the alignment at the fine-grained level is more significant than the coarse-grained level in terms of overall contribution. This also matches our intuition that building the fine-grained association between cognition and affection is closer to the conscious interaction process during human express empathy.

\subsection{Human Evaluation}

\begin{table}[t]
\centering
\resizebox{.95\columnwidth}{!}{
\begin{tabular}{l c c c c}
\hline
    Comparisons & Aspects & Win & Lose & $\kappa$ \\
\hline
    \multirow{3}{*}{\makecell[c]{CASE vs. EmpDG}}
    & Coh. & \textbf{48.1}$^\ddagger$ & 39.2 & 0.54 \\
    & Emp. & \textbf{51.9}$^\ddagger$ & 32.9 & 0.55 \\
    & Inf. & \textbf{58.9}$^\ddagger$ & 31.6 & 0.50 \\
\hline
    \multirow{3}{*}{\makecell[c]{CASE vs. KEMP}}
    & Coh. & \textbf{44.4}$^\dagger$ & 41.8 & 0.45 \\
    & Emp. & \textbf{50.0}$^\ddagger$ & 34.4 & 0.53 \\
    & Inf. & \textbf{51.1}$^\ddagger$ & 34.0 & 0.53 \\
\hline
    \multirow{3}{*}{\makecell[c]{CASE vs. CEM}}
    & Coh. & \textbf{45.9}$^\ddagger$ & 42.2 & 0.51 \\
    & Emp. & \textbf{53.2}$^\ddagger$ & 34.6 & 0.47 \\
    & Inf. & \textbf{57.8}$^\ddagger$ & 29.8 & 0.56 \\
\hline
\end{tabular}}
\caption{Human evaluation results (\%) of CASE and baselines. The agreement ratio kappa $\kappa \in [0.41,0.6]$ denotes the moderate agreement. $\dagger$,$\ddagger$ represent significant improvement with $p$-value $< 0.1/0.05$, respectively.}
\label{human}
\end{table}


\paragraph{Human Evaluation of CASE and Baselines}

Here, 200 contexts are randomly sampled and each context is associated with two responses generated from our CASE and baseline. Following \citet{sabour2021cem}, three crowdsourcing workers are asked to choose the better one (\textbf{Win}) from two responses by considering three aspects, respectively, i.e., (1) Coherence (\textbf{Coh.}): which model’s response is more fluent and context-related? (2) Empathy (\textbf{Emp.}): which model’s response expresses a better understanding of the user's situation and feelings? (3) Informativeness (\textbf{Inf.}): which model’s response incorporates more information related to the context? We use the Fleiss’ \textbf{kappa} ($\kappa$) \cite{fleiss1971measuring} to measure the inter-annotator agreement. As in Table \ref{human}, the results show that CASE outperforms three more competitive baselines on all three aspects. Especially, CASE outperforms baselines significantly in terms of empathy and informativeness, which shows the superior of modeling the interaction between cognition and affection of empathy, and supports the observations from automatic evaluation.

\paragraph{Human Evaluation on Variants of CASE}

To more intuitively verify the role of the key components of CASE in language expression, especially empathy ability, we conduct a scoring human evaluation for the variants of CASE. Besides the same settings as above, we require annotating the \textbf{Overall} preference score (1-5). As in Table \ref{human-ablation}, CASE achieves the highest scores in all aspects, indicating that all components contribute. The low empathy scores of “w/o Graph” and “w/o Align” as well as their variants further confirm the crucial role of graph structure and the effectiveness of alignment.

\begin{table}[t]
\centering
\resizebox{.95\columnwidth}{!}{
\begin{tabular}{l c c c c}
\hline
    Models & Coh. & Emp. & Inf. & Overall \\
\hline
    \textbf{CASE} & \textbf{3.88} & \textbf{3.48} & \textbf{3.62} & \textbf{3.58} \\
\hline
    w/o Graph & 3.72 & 3.02 & 3.51 & 3.38 \\
    w/o Align & 3.63 & 3.14 & 3.47 & 3.36 \\
\hline
    w/o CSGraph & 3.74 & 3.25 & 3.42 & 3.33 \\
    w/o ECGraph & 3.78 & 3.10 & 3.52 & 3.40 \\
    w/o CGAlign & 3.72 & 3.27 & 3.53 & 3.41 \\
    w/o FGAlign & 3.80 & 3.17 & 3.56 & 3.40 \\
\hline
\end{tabular}}
\caption{Human evaluation results of CASE's variants.}
\label{human-ablation}
\end{table}

\begin{table}[t]
\centering
\resizebox{.95\columnwidth}{!}{
\begin{tabular}{l c c c c}
\hline
    Models & PPL & Dist-1 & Dist-2 & Acc \\
\hline
    Bart & \textbf{15.17} & 2.77 & 16.41 & 0.419 \\
    BlenderBot & 15.22 & 2.70 & 16.20 & 0.470 \\
\hline
    \textbf{CASE-BlenderBot} & 15.40 & \textbf{2.92} & \textbf{17.66} & \textbf{0.492} \\
\hline
\end{tabular}}
\caption{Analysis of integrating pre-trained model.}
\label{pretrain}
\end{table}

\begin{table*}[t]
\centering
\resizebox{.95\linewidth}{!}{
\begin{tabular}{c | l}
\hline
    \textbf{Emotion} & \textbf{Terrified} \\
\hline
    \multirow{6}{*}{\makecell[c]{Context}} 
    & \textit{Speaker}: As a kid, I was so \textbf{frightened} any time there was a tornado warning. Not sure if it was the wizard \\
    & of oz effect or what i just remember being frozen with \textbf{fear}. \\
    & \textit{Listener}: Oh, wow! My son was \textbf{terrified} of tornadoes because of that darn movie, ...... \\
    & \textit{Speaker}: My mom said the \textbf{best} thing once when I was babysitting and a tornado warning went off and I did \\ & not want my own \textbf{fear} to affect the little boy I was with. She was on the phone and just said you have big \\ & plans for the future, right? Nothing is going to get in the way of that. For some reason, It worked a \textbf{charm}. \\
\hline 


    MoEL & I am sorry to hear that. I hope you can get better soon!  \\

    MIME & I am sure it was a great feeling.  \\

    EmpDG & I am glad you are ok! I am sure you will have a great time. \\

    KEMP & I would be too! I would have been terrified!   \\
    
    CEM & I am sure you will be fine! \\
\hline
    \textbf{\textit{Cognition}} & To make a plan, \textbf{To be safe}, To be helpful, \textbf{Have a good life} \\
    \textbf{\textit{Reaction}} & Scared, Hopeful, Nervous, \textbf{Good}, \textbf{Happy} \\
    \textbf{\textit{Concept}} & frighten, surprising, terrify, aghast, \textbf{good}, charming, \textbf{protect} \\
    \textbf{CASE} & That is \textbf{good}, I am \textbf{glad} you are able \textbf{to stay safe} and \textbf{get a little better}! \\
\hline
    \textbf{Ground-Truth} & Your mom sounds awesome! Wow, what a cool thing to say! \\
\hline
\hline
    \textbf{Emotion} & \textbf{Embarrassed} \\
\hline
    \multirow{4}{*}{\makecell[c]{Context}}
    & \textit{Speaker}: I burned my hair with my hair drier I am so embarrased to go out in public. \\
    & \textit{Listener}: I am \textbf{sorry} to hear that, my wife has \textbf{done} the same thing. She wore a hat for almost a month \\
    & when she left the house. \\
    & \textit{Speaker}: Yeah it is the \textbf{worst}, I look so \textbf{weird} with my hair like this it is so embarassing. \\
\hline
    
    
    MoEL & That is a pretty funny story. \\
    
    MIME & I am sure it was a good idea. \\
    
    EmpDG & Oh, that is a shame. I hope you are not too! \\
    
    KEMP & I am sure you will not have to be able to do it again.  \\
    
    CEM & I am sorry to hear that. \\
\hline
    \textbf{\textit{Cognition}} & \textbf{To look bad}, To get a haircut, \textbf{To fix it}, \textbf{To fix the hair}, To get a wig \\
    \textbf{\textit{Reaction}} &  Embarrassed, \textbf{Bad}, Ashamed, Sad \\
    \textbf{\textit{Concept}} & \textbf{bad}, ill, evil, freak, supernatural, exhausted, \textbf{hope}, sorrow \\
    \textbf{CASE} & I \textbf{hope} you are able to \textbf{get it fixed}, or just let her know \textbf{it is not too bad}.  \\
\hline
    \textbf{Ground-Truth} & Well, like I told her, it will grow out, it will just take time. \\
\hline
\end{tabular}}
\caption{Generated responses from baselines and CASE.
\textbf{Commonsense cognition}, \textbf{emotional reactions}, reasoned \textbf{emotional concepts} by \textbf{contextual words}, and corresponding \textbf{informative expressions} in responses are highlighted.}
\label{case}
\end{table*}

\subsection{Applicability Analysis}

To analyze the applicability of our method, we build it on the pre-trained model to explore whether it brings further benefits. We integrate BlenderBot \cite{DBLP:conf/eacl/RollerDGJWLXOSB21} into CASE by replacing the encoder and decoder, and take the vanilla Bart \cite{DBLP:conf/acl/LewisLGGMLSZ20} and BlenderBot as baselines. All pre-trained models are small versions. As in Table \ref{pretrain}, we found that CASE-BlenderBot integrating our method significantly outperforms finetune-only baselines. Although the overall performance of simple finetuning has achieved stage success, it is limited by the quality and scale of the dataset and lacks a more fine-grained design for the trait of human conversation. This also demonstrates the high-level applicability of our method for uncovering the underlying mechanisms of human conversation.

\subsection{Case Study}

Two cases from six models are selected in Table \ref{case}, among which CASE is more likely to express informative cognition in a highly empathetic tone. This is due to two main advantages:

(1) Effective alignment between cognition and affection on two levels. For example, in the first case, on the fine-grained level, CASE associates the cognition “\textit{to be safe}" with the affection “\textit{good}” (i.e., emotional reaction) to appease the user’s “\textit{Terrified}” experience, i.e., “\textit{to stay safe}” and “\textit{get a little better}”, in response. In the second case, on the coarse-grained level, in the user’s “\textit{Embarrassed}” emotional state, CASE expresses empathetic affection “\textit{it is not too bad}” with an informative cognitive statement, i.e., “\textit{get it fixed}”, in response.

(2) Accurate identification of the conversational emotion integrating emotional concepts and reactions, being consistent with “Acc”. For instance, in the first case, the correct conversational emotion “\textit{Terrified}” tends to be identified in the emotional concepts (“\textit{frighten, terrify, etc.}”), while in the second case, the one “\textit{Embarrassed}” tends to be identified in the emotional reactions (“\textit{embarrassed, ashamed, etc.}”). Compared with baselines that cannot correctly perform two cases simultaneously, CASE identifies correct emotion in both cases by integrating emotional concepts and reactions.

\section{Conclusion and Future Work}

In this paper, for responding empathetically, we propose CASE to align cognition and affection by simulating their conscious interaction in human conversation. Extensive experiments verify the superiority of CASE on overall quality and empathy performance. Our work will also encourage future work to model the more complex interaction between cognition and affection in human conversation as well as other human language behaviors \cite{liu-etal-2021-towards, zheng2023augesc}.

\section*{Limitations}

We discuss two limitations of this work as follows:

One limitation of our work is the lack of task-specific automatic metrics to evaluate the empathy of generated responses. Therefore, the evaluation of empathy relies more on human evaluation. Although human evaluation is a golden standard, automatic metrics help to conduct large-scale investigations. This is also a common limitation in current works on empathetic dialogue.

The second limitation is the passive response to the user's cognition and affection. In many scenarios, empathy is used as a strategy for emotional support by responding to the user's cognition and affection. However, besides passive response, emotional support also requires active emotion elicitation, which can be studied in future work.

\section*{Ethical Considerations}

In this paper, our experiments adopt the widely used \textsc{EmpatheticDialogues} benchmark, an open-source dataset collected from Amazon Mechanical Turk (MTurk) that does not contain personal information. We also ensure the anonymization of the human evaluation. We believe that this work honors the ethical code of ACL.

\section*{Acknowledgements}

This work was supported by the National Key Research and Development Program of China (No. 2021ZD0113304), the National Science Foundation for Distinguished Young Scholars (with No. 62125604) and the NSFC projects (Key project with No. 61936010).

This work was also supported by the National Natural Science Foundation of China (with No. 62272340, 61876128, 62276187).

\bibliography{anthology,EmpatheticDialog}
\bibliographystyle{acl_natbib}

\clearpage

\appendix

\section{Mutual Information Maximization}
\label{sec:mim}
Mutual information maximization (MIM) aims to measure the dependence between two random variables $X$ and $Y$, and the mutual information (MI) between them is defined as: 
$MI(X, Y)=D_{KL}(P(X, Y) \| P(X) P(Y))$. However, maximizing MI directly is normally intractable. A successful practice to estimate MI with a lower bound is InfoNCE \cite{DBLP:conf/iclr/KongdYLDY20}. Given two different views $x$ and $y$ of an input, InfoNCE is defined by:
\begin{equation}
\mathbb{E}_{P(X, Y)}[f_{\theta}(x, y)-\mathbb{E}_{Q(\widetilde{Y})}[\log \sum_{\widetilde{y} \in \widetilde{Y}} \exp f_{\theta}(x, \widetilde{y})]]+\log |\widetilde{Y}|,
\end{equation}
where $f_{\theta}$ is a learnable function with parameter $\theta$. The set $\widetilde{Y}$ draws samples from a proposal distribution $Q(\widetilde{Y})$, and it comprises $|\widetilde{Y}|-1$ negative samples and a positive sample $y$. One insight is that when $\widetilde{Y}$ always consists all values of $Y$ and they are uniformly distributed, maximizing InfoNCE is analogous to maximize cross-entropy loss:
\begin{equation}
\mathbb{E}_{P(X, Y)}[f_{\theta}(x, y)-\log \sum_{\widetilde{y} \in Y} \exp f_{\theta}(x, \widetilde{y})].
\end{equation}
It shows InfoNCE is relevant to maximize $P_\theta(y|x)$ and approximates summation over elements in $Y$ (i.e., partition function) by negative sampling \cite{DBLP:conf/cikm/ZhouWZZWZWW20, DBLP:conf/emnlp/Zhou0HZHHH22, DBLP:conf/coling/Zhou0YZHHH22}. Upon the formula, we replace $X$ and $Y$ with specific cognition and affection to maximize MI between them.

\end{document}